\documentclass[runningheads]{llncs}
\usepackage{graphicx}
\begin{document}
\title{Compression, The Fermi Paradox\\ and Artificial Super-Intelligence}
\titlerunning{Compression, The Fermi Paradox and Artificial Super-Intelligence}
%
\author{Michael Timothy Bennett}
\authorrunning{M. T. Bennett}
%
\institute{School of Computing, Australian National University, Canberra, Australia\\
\email{michael.bennett@anu.edu.au}}
\maketitle              
\begin{abstract}
The following briefly discusses possible difficulties in communication with and control of an AGI (artificial general intelligence), building upon an explanation of The Fermi Paradox and preceding work on symbol emergence and artificial general intelligence. The latter suggests that to infer what someone means, an agent constructs a rationale for the observed behaviour of others. Communication then requires two agents labour under similar compulsions and have similar experiences (construct similar solutions to similar tasks). Any non-human intelligence may construct solutions such that any rationale for their behaviour (and thus the meaning of their signals) is outside the scope of what a human is inclined to notice or comprehend. Further, the more compressed a signal, the closer it will appear to random noise. Another intelligence may possess the ability to compress information to the extent that, to us, their signals would appear indistinguishable from noise (an explanation for The Fermi Paradox). To facilitate predictive accuracy an AGI would tend to more compressed representations of the world, making any rationale for their behaviour more difficult to comprehend for the same reason. Communication with and control of an AGI may subsequently necessitate not only human-like compulsions and experiences, but imposed cognitive impairment.

\end{abstract}
\section{Introduction}
When examining what problems may arise in the pursuit of AGI, it may behoove us to consider explanations for The Fermi Paradox \cite{verendel_2017}, the contradiction between the apparent absence of extra-terrestrial life and its high probability. After all, both involve communication with a nonhuman intelligence. 

So, let us assume for the sake of argument that a non-human intelligence exists in our region of space, emitting signals in a similar medium to us (such as radio) neither attempting to contact nor hide from us; why might we have failed to identify or interpret the meaning of such signals, and what does this suggest for the pursuit of AGI?

\section{Symbolic Abstraction} 
First, let us consider what is necessary to infer the meaning of something. Natural language is a means of encoding and transmitting certain information between members of a species. What this information is, is debatable. A natural language model such as GPT-3 is trained only on text data \cite{floridi_2020}, implicitly endorsing the idea that meaning is just relations between words. While GPT-3 is capable of learning correlations in syntax to the extent that it can plausibly mimic human writing, it lacks any of the other sensorimotor information we might typically associate with words. Attempts to train models on multimodal sensorimotor data have yielded some success, with agents able to associate the sensory information of an object such as a cup with the signals that represent it \cite{b16,b10,b17}. Yet abstract notions such as ``politics" or ``ex-wife" would seem to require more than mere clustering of sensorimotor information.

One theory \cite{bennett_2021b} (the mirror symbol hypothesis), posits that the information encoded in natural language is not just sensorimotor stimuli but intent. Drawing on ideas from embodied and enactive cognition \cite{milkowski_2018,thompson_2007}, an organism's environment, sensors and actuators, the compulsions an organism labours under (such as hunger and pain) and so forth together specify an arbitrary task. Utility is replaced by a statement (a logical expression) characterizing sets of more or less desirable sensorimotor states (including the state of memory) - representing histories or situations from which plans and subsequently subgoals may be abducted. It is written in a physically implementable language such as arrangements of transistors or neurons, necessary and sufficient to reconstruct past experience given appropriate stimuli. If treated as a constraint to be satisfied, a solution or any subgoal derived thereof, expresses intent.  
Given an ostensive definition of a task (examples of successful task completion) there may be many apparently valid solutions, which vary in how well they generalise to unforeseen situations. The weaker and more general the solution, the closer it is to an idealised notion of intent (called an intensional solution).

In order to predict the intent of other agents, one agent assumes others conceive of the world as they do. It asks what subgoals might motivate the behaviour of other agents, given they are assumed to pursue a similar solution in general. It constructs a rationale for specific observed behaviour, to explain what another agent means to do (a subgoal, the pursuit of which would explain the observed behaviour). Subsequently, in order for communication to be possible two agents must possess approximately the same solution. Not only must they experience similar stimuli with which to construct symbolic abstractions, but imbue that stimuli with similar significance in terms of satisfying their compulsions. The solutions they construct then facilitate encoding and decoding of signals interpretable by both agents \cite{b7}. Any information not relevant to satisfying compulsions is not only meaningless but may be entirely ignored, which is consistent with observations of human behaviour \cite{zanto_2009,flowers_1979} (i.e. one may be unable to perceive something in the stream of sensorimotor stimuli because it is filtered out). 

This raises a few issues. The scope of a task is arbitrary, and so living as a typical human may be framed as such. Any nonhuman intelligence may face an entirely different task, to which they construct an entirely different solution. The difficulty this introduces is not only failure to understand what is meant, as in human language translation. 
Given two different solutions to two different tasks, stimuli may be imbued with either the same meaning, different meaning, or no meaning at all by one of those solutions (meaningful to one but not the other). The latter is particularly interesting, because such signals may not be recognised as intelligent behaviour (i.e. appear insane or mindless) or may perhaps be unnoticeable (i.e. the solution informs attention). 
In short, we may not realise something is a signal because what it conveys falls outside the scope of what humans are predisposed to notice or consider meaningful. While by definition we would be disinterested in such information, it may imply something which we would be consider meaningful (e.g. destroying the humans because of incomprehensible reasons). 

\section{Compression}
Allowing for the above, one may still mechanically assess information content and decode signals \cite{mackay_2003}, to glean something of potential meanings by correlation, even if they are incomprehensible.
However, the same information can be represented in many different ways, compressed to different extents.
As the volume of digital information exchanged and stored by humans each day has increased, so has the utility of compression. Streaming services such as Youtube make extensive use of compression to reduce the cost of transmission and storage. Another intelligence may also wish to reduce the cost of transmission and storage by employing the most effective compression they possess. 
Taking into account the decoder which reconstructs a signal, the greatest extent to which a signal may be compressed is its Kolmogorov Complexity \cite{kolmogorov_1963}; the length of the smallest self extracting archive capable of reproducing that signal. Such a compressed signal contains no discernible pattern of which we might take advantage to more efficiently represent the signal without discarding information. 
Uniformly distributed, random noise is also not compressible. There is no pattern. A highly compressed signal may appear to be nothing more than random noise to any observer lacking the appropriate decoder. Any advanced intelligence may compress information to the extent that we mistake their signals for noise \cite{Gurzadyan_2016}.

The ability to generalise is closely related to compression \cite{solomonoff_1964a,solomonoff_1964b}, with more compressed representations yielding better accuracy \cite{b27} for the same reason that there is only one straight line interpolating any two points, but infinitely many polynomials of higher degree. As stated earlier, solutions to a task may vary in how well they generalise, with an intensional solution being the most general. 
A super-intelligent AGI would construct such a solution \cite{bennett_2021b}, which is likely to be among the most compressed \cite{b27}, meaning any rationale for its decisions may be uninterpretable for the same reason a highly compressed signal is. Subsequently the onus is on the more intelligent agent to communicate in terms that the a less intelligent agent comprehends. Given human-like sensors, actuators, compulsions and so forth, situated in a human environment, an AGI may construct a solution similar enough to humans to facilitate communication. However, to guarantee interpretability and control of an AGI may require restricting how it constructs solutions such that it is, on some level, cognitively impaired \cite{Trazzi_2018}.

\end{document}